\documentclass[10pt,twocolumn,letterpaper]{article}

\usepackage{wacv}
\usepackage{times}
\usepackage{epsfig}
\usepackage{graphicx}
\usepackage{amsmath}
\usepackage{amssymb}
\usepackage{booktabs}
\usepackage{float}
\usepackage{longtable}

\wacvfinalcopy 

\ifwacvfinal
\usepackage[breaklinks=true,bookmarks=false]{hyperref}
\else
\usepackage[pagebackref=true,breaklinks=true,colorlinks,bookmarks=false]{hyperref}
\fi

\begin{document}

\title{Match Cutting: Finding Cuts with Smooth Visual Transitions}

\author{
    Boris Chen \qquad Amir Ziai \qquad Rebecca S. Tucker \qquad Yuchen Xie \\
    \{bchen, aziai, btucker, yxie\}@netflix.com \\
	Netflix Inc. \\ Los Gatos, CA, USA
}

\maketitle
\thispagestyle{empty}

\begin{abstract}
A match cut is a transition between a pair of shots that uses similar framing, composition, or action to fluidly bring the viewer from one scene to the next. Match cuts are frequently used in film, television, and advertising. However, finding shots that work together is a highly manual and time-consuming process that can take days. We propose a modular and flexible system to efficiently find high-quality match cut candidates starting from millions of shot pairs. We annotate and release a dataset of approximately 20k labeled pairs that we use to evaluate our system, using both classification and metric learning approaches that leverage a variety of image, video, audio, and audio-visual feature extractors. In addition, we release code and embeddings for reproducing our experiments at \href{http://github.com/netflix/matchcut}{github.com/netflix/matchcut}.
\end{abstract}


\begin{figure}
    \centering
    \includegraphics[width=1.0\linewidth]{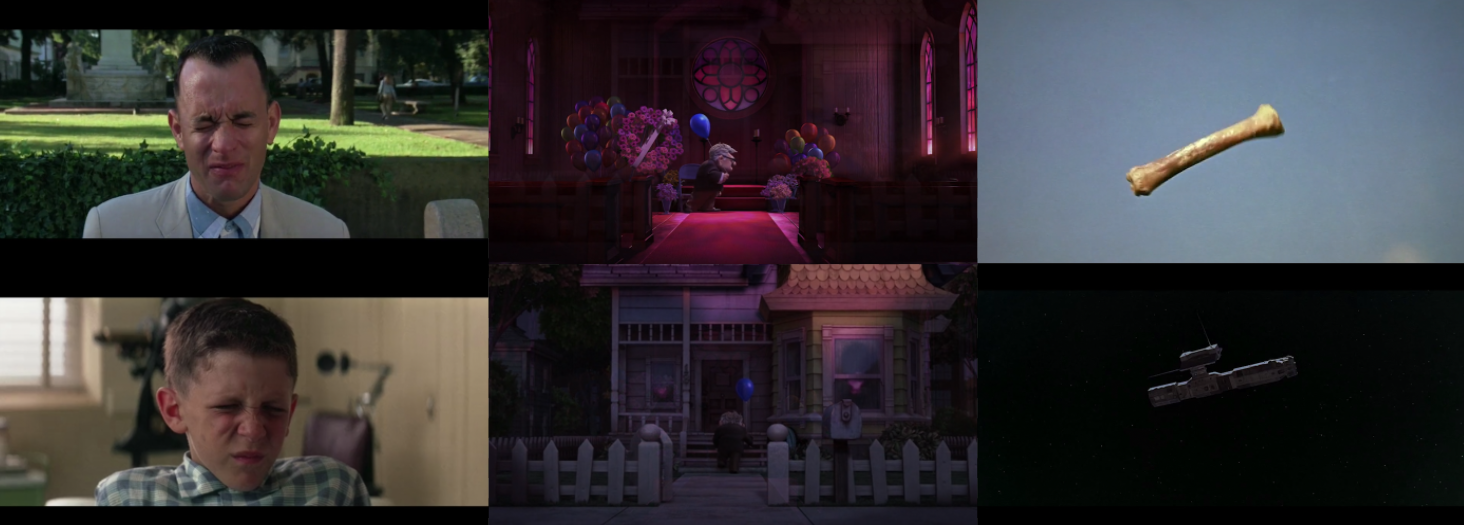}
    \caption[]{Three example match cuts where the framing of the subject is matched: (left) Forrest Gump (1994), (center) Up (2009), and (right) 2001: A Space Odyssey (1968).}
    \label{fig:frame_match_cuts}
\end{figure}

\section{Introduction}
In film, a shot is a series of frames representing an uninterrupted period of time between two cuts \cite{braudy1995film}. A match cut is a transition between a pair of shots that uses similar framing, composition, or action to fluidly bring the viewer from one scene to the next. It is a powerful visual storytelling tool used to create a connection between two scenes.

For example, a match cut from a person to their younger or older self is commonly used in film to signify a flashback or flash-forward to help build the backstory of a character. Two example films that used this are Forrest Gump (1994) \cite{forrest_gump_1994} and Up (2009) \cite{up_2013} (Fig. \ref{fig:frame_match_cuts}). Without this technique, a narrator or character might have to explicitly verbalize that information, which may ruin the flow of the film.

A famous example from Stanley Kubrik's 2001: A Space Odyssey \cite{kubrick_clarke_1968} is also shown in Fig. \ref{fig:frame_match_cuts}. This iconic match cut from a spinning bone to a spaceship instantaneously takes the viewer forward millions of years into the future. It is a highly artistic edit which suggests that mankind's evolution from primates to space technology is natural and inevitable.

\begin{figure}
    \centering
    \includegraphics[width=0.98 \linewidth]{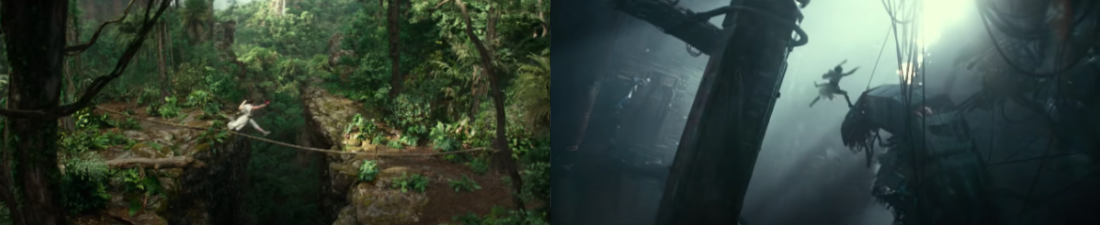}
    \caption{A match cut from the \emph{Star Wars: The Rise of Skywalker} (2019) \cite{starwars_2019} trailer. The trailer editor took two shots from the different scenes with similar jump motions and cut them together. The matched motion gives the illusion of one continuous jump.}
    \label{fig:starwars_matchcut}
\end{figure}

Match cuts can use any combination of elements, such as framing, motion, action, subject matter, audio, lighting, and color. In this paper, we will specifically address two types: (1) character frame match cuts, in which the framing of the character in the first shot aligns with the character in the second shot, and (2) motion match cuts, where shots are matched together on the basis of general movement. Motion match cuts can use common camera movement (pan left/right, zoom in/out) or motion of subjects. They create the feeling of smooth transitions between inherently discontinuous shots. An example is shown in Fig. \ref{fig:starwars_matchcut}.

Match cutting is considered one of the most difficult video editing techniques \cite{douglass1996art}, because finding a pair of shots that match well is tedious
and time-consuming. For a feature film, there are approximately 2k shots on average, which translates to 2M possible shot pairs, the vast majority of which will not be good match cuts. An editor typically watches one or more long-form videos and relies on memory or manual tagging to identify shots that would match to a reference shot observed earlier. Given the large number of shot pairs that need to be compared, it is easy to overlook many desirable match cuts.


Our goal is to make finding match cuts vastly more efficient by presenting a ranked list of match cut pair candidates to the editors, so they are selecting from, e.g., the top 50 shot pairs most likely to be good match cuts, rather than millions of random ones. This is a challenging video editing task that requires complex understanding of visual composition, motion, action, and sound. 

Our contributions in this paper are the following:
(1) We propose a modular and flexible system for generating match cut candidates. Our system has been successfully utilized by editors in creating promotional media assets (e.g. trailers) and can also be used in post-production to find matched shots in large amount of pre-final video. (2) We release a dataset of roughly 20k labeled match cut pairs for two types of match cuts: character framing and motion. (3) We evaluate our system using classification and metric learning approaches that leverage a variety of image, video, audio, and audio-visual feature extractors. (4) We release code and embeddings for reproducing our experiments.


\section{Related Work}
\noindent \textbf{Computational video editing}
There is no computational or algorithmic approach to video editing that matches the skill and creative vision of a professional editor. However, a number of methods and techniques have been proposed to address sub-problems within video editing, particularly the automation of slow and manual tasks.


Automated video editing techniques for specialized non-fiction videos has seen success with rules-based methods, such as those for group meetings \cite{meeting_capture,multi_party}, educational lectures \cite{HWG07}, interviews \cite{berthouzoz2012} and social gatherings \cite{multiple_social_cameras}. Broadly speaking, these methods combine general film editing conventions (e.g. the speaker should be shown on camera) with heuristics specific to the subject domain (e.g. for educational lectures, the white board should be visible). 

Computational video editing for fictional works tends to fall in one of two lines of research: transcript-based approaches \cite{cve_dialogue,wav,fried2019textbased,QuickCut} and learning-based approaches \cite{Pardo_2021_ICCV}. Leake et. al. \cite{cve_dialogue} generates edited video sequences using a standard film script and multiple takes of the scene, but their work is specific to dialogue-driven scenes. Two similar concepts, Write-A-Video \cite{wav} and QuickCut \cite{QuickCut}, generate video montages using a combination of text and a video library. Learning-based approaches have seen success in recent years, notably in Learning to Cut \cite{Pardo_2021_ICCV}, which proposes a method to rank realistic cuts via contrastive learning \cite{chopra2005learning}. The MovieCuts dataset \cite{pardo2021moviecuts} includes match cuts as a subtype, though it is by far the smallest category and does not distinguish between kinds of match cuts. In contrast, we release a data set of 20k pairs that differentiate between frame and motion cuts, with the goal of finding these pairs from shots throughout the film instead of detecting existing cuts. Our work advances learning-based computational video editing by introducing a method to generate and then rank proposed pairs of match cuts without
fixed rules or transcripts.


\noindent
\textbf{Video Representation Learning} Self-supervised methods have dominated much of the progress in multi-modal media understanding in recent years \cite{yan2022multiview,liu2021swin,feichtenhofer2019slowfast,jing2020self}. CLIP \cite{openai-clip} was an early example of achieving impressive zero-shot visual classification following self-supervised training with over 400M image-caption pairs. Similar advances have been made for audio \cite{guzhov2021audioclip} and video \cite{CLIP4Clip} by utilizing different augmented views of the same modality \cite{chen2021exploring,chen2020simple,grill2020bootstrap,recasens2021broaden}, or by learning joint embeddings of short \cite{guzhov2021audioclip,alwassel2020self,CLIP4Clip} or long-form \cite{kalayeh2021watching} videos. Our system leverages such work for learning video representations that capture matching video pairs for the task of match cutting.


\noindent
\textbf{Movie understanding}
There is a deep and rich literature on models that understand and analyze the information in movies. Many movie-specific datasets \cite{huang2020movie} have been developed that have enabled research into a variety of topics such as human-centric situations \cite{vicol2018moviegraphs}, story-based retrieval \cite{bain2020condensed}, shot type classification \cite{rao2020unified}, narrative understanding \cite{bain2020condensed,bhat2021hierarchical,lapata2021movie}, and trailer analysis \cite{huang2018trailers}. We release a dataset which contributes a novel and challenging movie understanding task.


\section{Methodology}\label{sec:methodology}
\begin{figure*}
    \centering
    \includegraphics[width=0.9\linewidth]{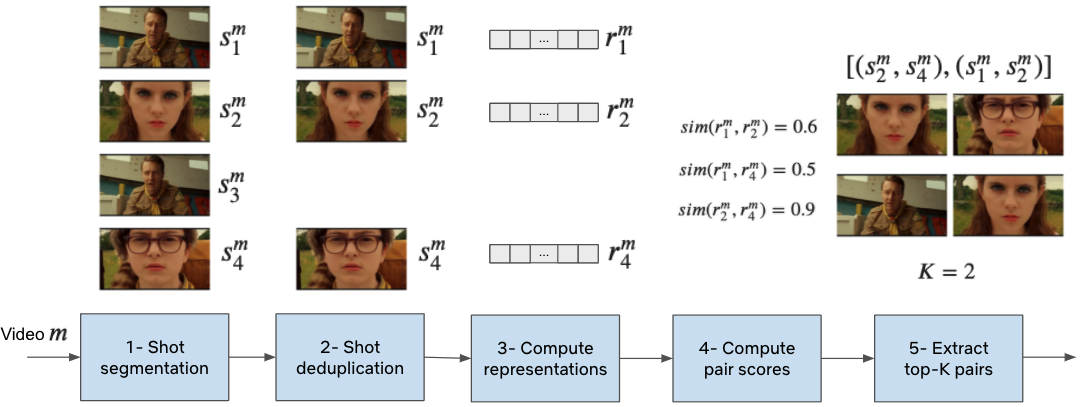}
    \caption{System diagram for generating candidate match cut pairs. The input is a video file for movie $m$ and the output is $K$ match cut candidates. (1) Video is split into shots using a shot segmentation algorithm. (2) Near-duplicate shots are removed. (3) A tensor representation $r^m_i$ is computed for each shot $s^m_i$ using an encoder. (4) All unique shot pairs are enumerated and a score function $sim$ is used to compute the similarity between shot representations. (5) The top-$K$ pairs with highest similarity are returned. We show an illustrative example with four shots from Moonrise Kingdom (2012) \cite{anderson_2012} and $K=2$. }
    \label{fig:system}
\end{figure*}

In this section, we present a flexible and scalable system for finding $K$ matching shot pairs given a video. This system consists of five steps, as depicted in Fig. \ref{fig:system}.

\subsection{Preprocessing}\label{sec:preprocessing}
The first two steps of our system segment a video into a sequence of contiguous and non-overlapping shots and remove near-duplicate shots. Although we present concrete implementations for these steps, our system is agnostic to these choices.

\textbf{Step 1: Shot segmentation.}
For each movie $m$, we run a shot segmentation algorithm to split that title into $n_m$ shots. Let $S^m = \{s^m_i \}_{i=1}^{n_m}$ be the set of shots where $s^m_i$ corresponds to the $i$-th shot of the $m$-th movie. Shot $s^m_i$ consists of an ordered set of frames $F^m_i = \{f_{(i,j)}^m\}_{j=1}^{l^m_i}$, where $f_{(i,j)}^m$ is the j-th frame of $s^m_i$, and $l^m_i$ is the number of frames in $s^m_i$. We use a custom shot segmentation algorithm but similar results can by achieved with PySceneDetect \cite{pyscene} or TransNetV2 \cite{soucek2020transnetv2}.

\textbf{Step 2: Near-duplicate shot deduplication.}
Matching shots should have at least one difference in character, background, clothing, or character age. Therefore, we remove near-duplicate shots (e.g. two shots of the same character in the same scene and framing, but with a slightly different facial expression).

Our specific methodology for deduplication is as follows:
we first extract the center frame $c^m_i$ for each shot $s^m_i$ defined as $c^m_i = f_{(i, \lfloor l^m_i / 2 \rfloor)}^m$. For each center frame, we extract the penultimate embeddings out of MobileNet \cite{MobileNets} pretrained on ImageNet \cite{krizhevsky2012imagenet}. Let $e^m_i = \text{enc}(c^m_i) \in 	\mathbb{R}^{1024}$ be the embedding for frame $c^m_i$ where $\text{enc}$ takes an image and outputs a 1024-dimensional vector.

We define the set of duplicate shot indices for movie $m$ as 
\begin{equation}
 D^m = \{j|  i,j \in \{1,2,\ldots,n_m\}, i < j, \text{cos}(e^m_i, e^m_j) \ge T_d \}   
\end{equation}
where $\text{cos}$ computes the cosine similarity between a pair of embeddings and $T_d$ is the similarity threshold.

Finally, the set of deduplicated shots for movie $m$ can be constructed by excluding the shots corresponding to the indices in $D^m$ as follows: $S^m_d = \{s^m_i | i \in \{1,2,\ldots,n_m\}, i \not \in D^m\}$. We leverage the \texttt{imagededup} \cite{idealods2019imagededup} library and find that setting $T_d = 0.8$ removes most of the near-duplicates. 

\subsection{Shot Pair Ranking}\label{sec:shot_rep}
Steps 3-5 score and rank pairs of deduplicated shots following step 2.

\textbf{Step 3: Shot representation computation.} 
In this step, we compute a tensor representation $r^m_i$ for each shot $s^m_i$. Representations for different shots need to preserve some notion of similarity for matching pairs. Representations can be extracted using any video, image, audio, text, or multi-modal encoders. We present a few such choices in the upcoming sections.

\textbf{Step 4: Shot pair score computation.}
In this step, we enumerate all unique shot pairs for movie $m$,
\begin{equation}
P^m = \{ (s^m_i, s^m_j) | s^m_i, s^m_j \in S^m_d, i < j \}
\end{equation}
and compute a similarity score $\text{sim}(r^m_i, r^m_j) \in \mathbb{R}$ for each pair of shots $(s^m_i, s^m_j)$. This similarity score is used for ranking pairs where higher-scoring pairs are considered higher quality. The function $\text{sim}$ can be any function that takes a pair of tensors and outputs a real scalar. This function can be chosen beforehand (e.g. cosine similarity) or learned through supervision.

\textbf{Step 5: Top-$K$ pair extraction.}
This step simply ranks the results from the previous step and returns the top-$K$ pairs.

\subsection{Heuristics}\label{sec:heuristics}

We define a heuristic $h$ as a specific combination of shot representation and predetermined scoring function. These heuristics serve two functions. We use them to \emph{generate} candidate pairs for manual annotation by video editors, and then also to \emph{evaluate} the annotated data set. Here, evaluate means that we use the heuristic to rank the candidate pairs and compute the average precision of that ranked list. More details about evaluation can be found in Supplementary \ref{supp:evaluation}.

We leverage four of the heuristics presented in this section ($h_1$, $h_2$, $h_4$, and $h_5$) to generate candidate pairs for annotation in Sec. \ref{sec:data} and report how all of the heuristics perform on our dataset in Sec. \ref{sec:experiments}.

\textbf{Heuristic 1 ($h_1$): equal number of faces.}
One very crude heuristic for character frame match cutting is to consider pairs where the number of faces between the two shots is equal. For the shot representation, we extract the center frame and use a face detection model (Inception-ResNet-v1 \cite{szegedy2017inception} pretrained on VGGFace2 \cite{cao2018vggface2}) to determine the number of faces. The scoring function outputs 1 if the two shots have the same number of faces, and 0 otherwise.

\textbf{Heuristics 2 ($h_2$) and 3 ($h_3$): Instance segmentation.}


These heuristics are designed for character frame matching. We leverage instance segmentation to extract pixel-level representations of the presence of people in a frame. In other words, we can extract the silhouette of characters, which contain rich information about the character’s framing in the image.

The instance segmentation model takes a single center frame $c^m_i$ and returns a set of instance-level binary masks $B^m_i = \{ b^m_{(i, x)} \}_{x=1}^{u^m_i}$, where $u^m_i$ is the number of such instances, $b^m_{(i, x)} \in \{0, 1\}^{W \times H}$ is the binary mask of the $x$-th instance, $W$ is the width of the binary mask, and $H$ is the height (same shape as $c^m_i$).

We use the union of all binary masks $b^m_i = \bigcup_{x=1}^{u^m_i}b^m_{(i, x)}$ as the shot representation $r^m_i$ for heuristic 2 $(h_2)$. Intuitively, a perfect character match cut will involve characters with the exact same binary mask. The Intersection over Union ($IoU$) metric captures this notion well and we use it as the similarity score for $h_2$. Concretely, $IoU$ takes two binary masks and returns a real value between 0 and 1, which captures how well the masks overlap (the higher the better):
\begin{equation}
IoU(b^m_i, b^m_j) = | b^m_i \cap b^m_j | / | b^m_i \cup b^m_j |
\end{equation}
where $b^m_i, b^m_j$ are binary masks for center frame $c^m_i$ and $c^m_j$, $b^m_i \cap b^m_j$ is the set of person pixels that are shared between the two masks, and $b^m_i \cup b^m_j$ is the set of pixels that contain a person pixel in at least one of the masks. If either mask is empty, we set $IoU = 0$.

\begin{figure}
    \centering
    \includegraphics[width=1\linewidth]{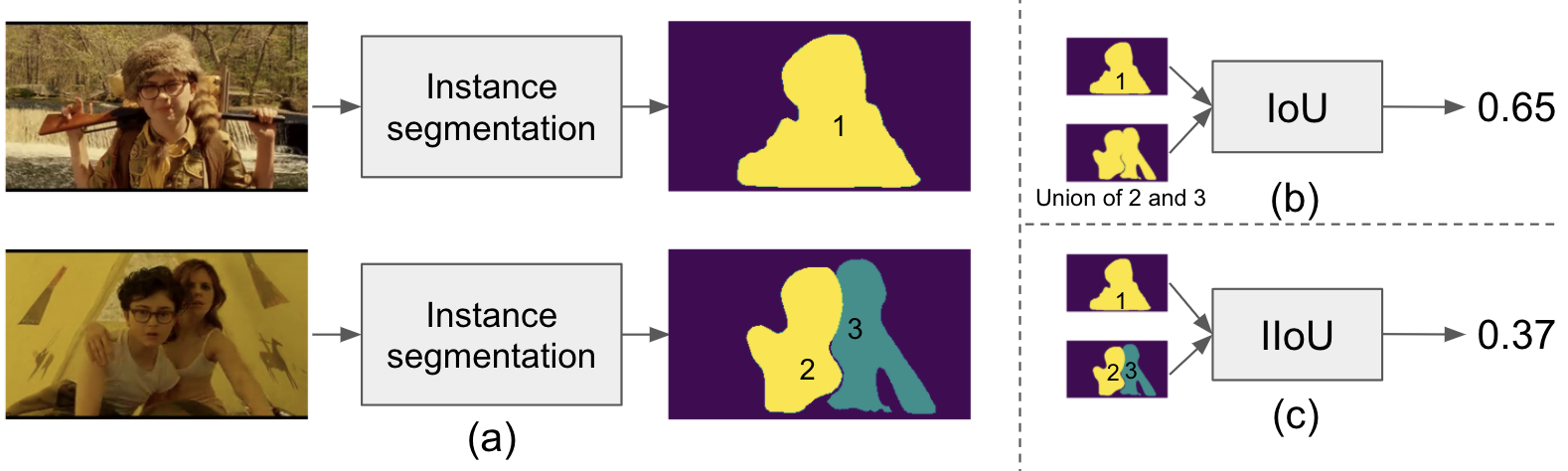}
    \caption{(a) Two frames from the Moonrise Kingdom \cite{anderson_2012} are passed through an instance segmentation network to obtain instance-level binary masks. (b) For $h_2$ we use the union of masks as the representation and IoU as the similarity score function. (c) for $h_3$ we preserve the instance-level binary masks for the representations, and use IIoU as the similarity score function.}
    \label{fig:instance_seg}
\end{figure}

Taking the union of instance-level binary masks can lead to matching the wrong number of characters between two shots, as depicted in Fig. \ref{fig:instance_seg}. Instead, for heuristic 3 ($h_3$) we use $B^m_i$ as the representation of $s^m_i$ and use Instance IoU as the similarity score function.

We start by associating character instances across center frames, which ensures that each instance in $B^m_i$ is matched to at most one instance in $B^m_j$. Inspired by SORT \cite{Bewley2016_sort}, we formulate this association as a linear assignment problem with assignment cost computed using the negative IoU of instance masks. We denote $A^m_{(i, j)}$ as the set of associated instance mask pairs between $c^m_i$ and $c^m_j$. The instance-level IoU (IIoU) is computed as follows:
\begin{equation}
IIoU(B^m_i, B^m_j) = \dfrac{ \sum_{ (b^m_{(i, x)}, b^m_{(j, y)}) \in A^m_{(i,j)} }{ |b^m_{(i, x)} \cap b^m_{(j, y)}| } }{ |b^m_i \cup b^m_j| }
\end{equation}


Even with this improvement, both of these heuristics fail in at least two scenarios. First, segmentation inaccuracies can lead to false positives or false negatives. Second, two frames with very different character poses sometimes produce highly similar binary masks (e.g. matching the face of one character to the back of another).


We use the PyTorch \cite{paszke2019pytorch} implementation of Mask R-CNN with a ResNet-50-FPN backbone \cite{resnet,Yakubovskiy:2019}, pretrained on COCO train2017 \cite{coco}, filter out all instance types except for ``person", and use a 0.5 threshold.

\textbf{Heuristics 4 ($h_4$) and 5 ($h_5$): Optical Flow.}
We use heuristics 4 and 5 to generate annotation candidates for motion match cutting. The editors are looking for pairs of shots with similar motions so that they can edit together to create the continuation. The criterion for good motion match cuts is a similarity in the movement of the camera or subjects between the shots. Because movement is so critical to this kind of match cut, we cannot simply take the static center frame of each shot as we do for heuristics 1-3.

Instead, we need a way to quantify the motion of the shot across all frames. Optical flow refers to the task of estimating the movement of boundaries, edges, and objects within a video or sequence of ordered images. Dense optical flow is the task of estimating the movement of each pixel within a video frame. Sparse optical flow, on the other hand, only tracks the movement of key points within the frame. Optical flow has been traditionally formulated as an optimization problem \cite{opticalflow_opt1,opticalflow_opt2,opticalflow_opt3} or as a gradient-based estimation. However, in recent years, deep-learning based models have become a viable alternative to these methods.

Both heuristics for motion match cuts use the following general procedure. For each sequential pair of frames in the shot $s^m_i$, we compute the optical flow, which yields a tensor of size $W\times H \times2$, representing the horizontal and vertical motion for each pixel. Here $W$ and $H$ are the width and height, respectively, of the frame in pixels. The optical flow tensor for consecutive frames $f^m_{(i, j)}$ and $f^m_{(i, j+1)}$ is $q^m_{(i,j)} = OF(f^m_{(i, j)}, f^m_{(i, j+1)})$, where $OF$ is the specific choice of optical flow implementation.

\begin{figure}
    \centering
    \includegraphics[width=0.49\linewidth]{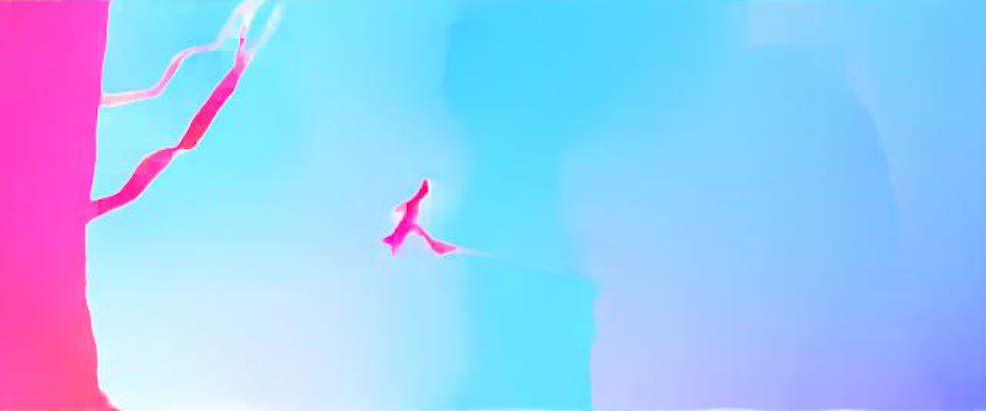}
    \includegraphics[width=0.49\linewidth]{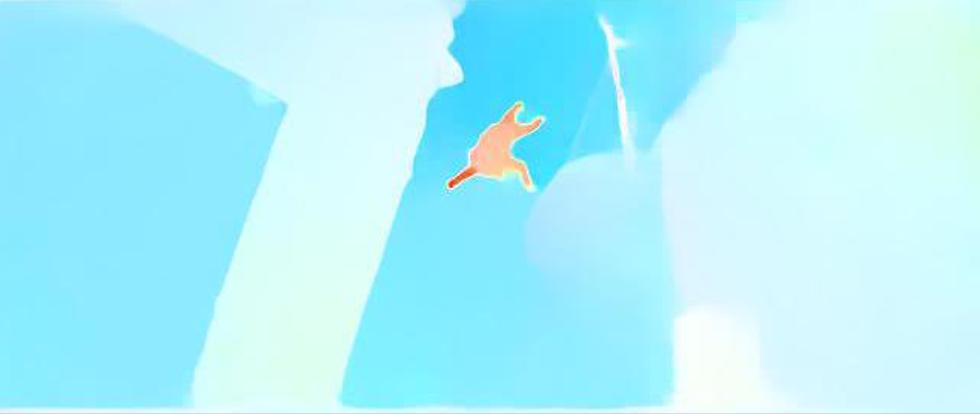}
    \caption{An optical flow result for the match cut in Fig. \ref{fig:starwars_matchcut} from the trailer for \emph{Stars Wars: The Rise of Skywalker} \cite{starwars_2019}. In this visualization, the color represents the direction of motion. Pink pixels are moving to the right and blue pixels are moving to the left. The intensity of the color represents the magnitude of the motion.}
    \label{fig:starwars_optical_flow}
\end{figure}

For each shot, we average the optical flow tensors across all the frames in the shot $s^m_i$:

\begin{equation}
    Q^m_i = \left( \sum_{j=1}^{l^m_i - 1} q^m_{(i, j)} \right) / (l^m_i - 1)
\end{equation}
This allows us to compare shots of different length. An example optical flow output can be seen in Fig. \ref{fig:starwars_optical_flow}.

For heuristic 4 ($h_4$), we use the \texttt{opencv} implementation \cite{opencv_library} of the Farnebäck method to compute dense optical flow \cite{two-framemotion}. This method approximates the neighborhood of each pixel with a quadratic polynomial and uses differences in the polynomials to estimate the per-pixel displacement. 

For heuristic 5 ($h_5$), we use a deep-learning-based method: Recurrent All-Pairs Field Transforms (RAFT) \cite{RAFT}. Its network architecture comprises three components: a feature extractor, a correlation layer that creates a 4D correlation volume for all pixel pairs, and a recurrent GRU-based update operator. We are using the version of the model that has been pretrained on the FlyingThings \cite{flyingthings} dataset. The optical flow visualizations in Fig. \ref{fig:starwars_optical_flow} were generated using RAFT.

We use $r^m_i = flatten(Q^m_i)$ and cosine similarity as the similarity function for this heuristic. In practice, we sample every four frames to save computation and achieve similar results to using every frame.

\subsection{Scalability}\label{scalability}
Our proposed system can be used to find match cuts both within a title and across multiple titles. However, the number of shot pairs that need to be compared is quadratic in the number of shots, which can quickly become a bottleneck if we want to find match cuts across multiple titles. To avoid this, we can replace steps 4 and 5 in Section \ref{sec:shot_rep} with an approximate nearest neighbors approach (ANN) such as FAISS \cite{johnson2019billion}. For this approach, we first need to build an index of the representations that we computed in Section \ref{sec:shot_rep}. Once this index is built, we can retrieve the top-$K$ results for each shot, and then compute the global top-$K$ pairs.

Although this approach is significantly more efficient, it presents three issues. First, the approximate nature of ANN methods may lead to imprecise results. Second, ANN methods don't support arbitrary functions for computing nearest neighbors. Third, representations must have the same shape. We explore this trade-off further in Sec. \ref{sec:experiments}.
\section{Data}\label{sec:data}

The dataset we release with this paper contains $\sim$20k pairs of labeled shots. We include seven embeddings for each shot (described in Sec. \ref{sec:experiments}), shot boundary timestamps, and the annotated labels. In this section, we describe the process for the collection of this dataset.



\subsection{Movie Set Selection and Pre-processing}
Here we select a set of movies, segment these movies into shots, remove near-duplicate shots, and construct the set of all shot pairs within the same movie (intra-movie).

\textbf{Movie set selection}
We selected 100 movies from the MovieNet \cite{huang2020movie} dataset which are diversified across genre, release year, and country of origin. The full list of movies can be found in Supplementary \ref{supp:title}.

\textbf{Shot segmentation}
We segmented each of the 100 movies into shots as described in Sec. \ref{sec:preprocessing}. Following this step, we are left with 128k shots across all movies, which translates into over 8.2B unique shot pairs that could be considered for annotation.

\textbf{Shot deduplication}
We use the deduplication methodology described in Sec. \ref{sec:preprocessing} to remove near-duplicate shots. This step shrinks the overall number of shots by over 40\%, from 128k to 75k, which also shrinks the number of pairs to 2.8B. 

\textbf{Limit to intra-movie matches}
Though inter-movie match cuts are also interesting, for this study we only consider intra-movie pairs--which leaves $\sim$35M shot pairs. The vast majority of these pairs are not match cuts, so we employ heuristics to further reduce the annotation candidate set and increase the likelihood of finding positive pairs relative to random sampling.

\subsection{Annotation Candidate Pair Generation} \label{sec:data_heuristics}
Using the heuristics discussed in Sec. \ref{sec:heuristics}, we score, rank, and retrieve the top 50 shot pairs for each movie $m$, which we denote $P^m_{h_i}$. Note that $P^m_{h_i} \subseteq P^m$ and that $|P^m_{h_i}|=50$. For each type we utilized two heuristics $h_i$ and $h_j$, and used the union of the resultant sets as the dataset for annotation.

We use $h_1$ and $h_2$ for character frame match cutting and $h_4$ and $h_5$ for motion match cutting. (Heuristic $h_3$ was not available during the annotation process.) In other words, $P^m_{h_1} \cup P^m_{h_2}$ is the set of pairs that we annotated for character frame, and $P^m_{h_4} \cup P^m_{h_5}$ for motion.



\begin{figure}
    \centering
    \includegraphics[width=0.98\linewidth]{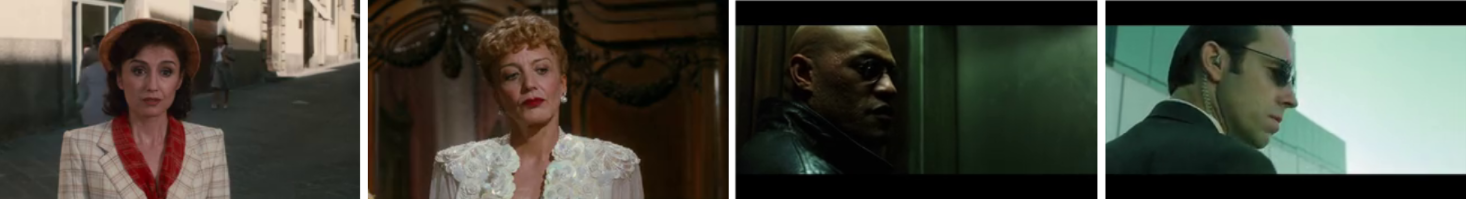}
    \caption{(left) A frame match cut from \emph{Life is Beautiful} (1997) \cite{benigni_1997}, (right) A frame match cut from \emph{The Matrix} (1999) \cite{wachowski_1999}.}
    \label{fig:lib_matchcut}
\end{figure}
\begin{figure}
    \centering
    \includegraphics[width=0.98\linewidth]{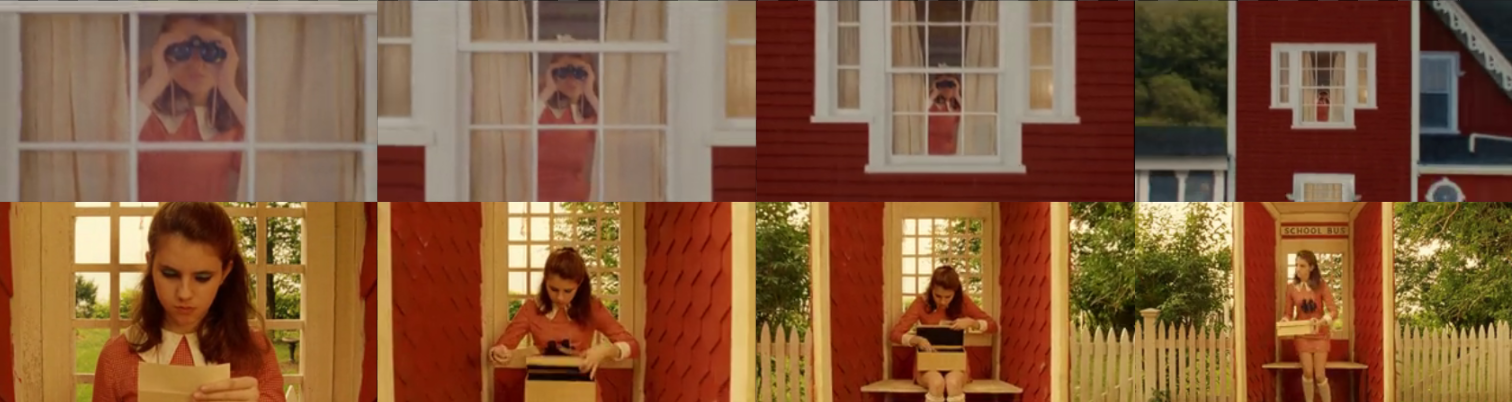}
    \caption{A motion match cut from \emph{Moonrise Kingdom} (2012) \cite{anderson_2012} where the primary motion detected by optical flow is from the camera zooming out.}
    \label{fig:mk_matchcut}
\end{figure}

\textbf{Heuristic-discovered match cuts }
We present a few examples from our match cutting heuristics. Fig. \ref{fig:lib_matchcut} contains examples of character frame match cut $h_2$. Fig. \ref{fig:mk_matchcut} is an example of a motion match cut found by $h_5$. 



\subsection{Data Collection}
\subsubsection{Task Definitions}
Our data collection process started with inspecting and refining the definitions of each match cutting type with a group of three senior in-house video editors. After arriving at consistent definitions, we developed reference material, which included illustrative examples of positive and negative pairs.

Our senior video editors trained two separate sets of three annotators, one for each type. Due to the technical nature of this task, we selected video editors at a video editing agency that we had previously worked with. We asked annotators to assign a binary label to each pair they were presented with. The training involved an hour-long walk-through of guidelines and examples.

\subsubsection{Annotations}\label{annotations}
We used six annotators: three for character frame match cuts and three for motion match cuts, so each candidate pair was annotated by three annotators. For frame match cuts, our annotators labeled 9,985 pairs and were in perfect agreement (all 3 annotators chose the same label) for 84\% of the pairs. For motion match cuts, our annotators classified 9,320 pairs and were in perfect agreement for 75\% of them. We attribute the lower annotator agreement for motion match cutting to the fact that it is a more difficult, subjective, and nuanced task. We use the majority vote for each pair as the final label for the rest of this paper. However, annotator-level data is also made available in the dataset. Roughly 8.7\% of frame match pairs were majority labeled positive, and 9.9\% for motion.



\subsubsection{Random negative pairs}
Since we used heuristics to generate annotation candidates, our dataset may not reflect many pairs that the models are likely to encounter. For instance, we only used shots with faces of people for frame match cutting, but our system needs to be able to score pairs that contain no faces. To reduce this bias, we randomly sample 50 pairs for each title and append these pairs as negative examples to the dataset we use for training and evaluating models in Sec.  \ref{sec:experiments}. These pairs have no overlap with the annotated pairs and, since they were drawn at random, they are extremely unlikely to contain positive pairs. More information about the dataset and its associated statistics can be found in Supp. \ref{supp:annotation}.

\section{Experiments} \label{sec:experiments}


Our experiments explore two goals: finding a scalable solution to a quadratic problem and high retrieval quality. 
The heuristic-based candidate generators are useful to aid labeling by suggesting pairs for annotation, but a learned model can perform better than heuristics. 
High model accuracy translates to saved time and higher quality cuts for our video editors.

Experiment 1 explores a binary classification setup, which gives us high retrieval quality, but at the expense of scalability, since every shot pair will need to be run through the classifier. 
Experiment 2 uses metric learning to provide both high retrieval quality and scalability.
For both sets of experiments, the shot segmentation and deduplication steps remain fixed.



\subsection{Setup}

We split our dataset at the movie level by randomly selecting 60, 20, and 20 titles for train, validation, and test sets respectively. For all experiments we train 5 models using different seeds, compute average precision ($AP$) on the validation dataset ($AP_{val}$) for each model, and finally report test $AP$ ($AP_{test}$) for the model with the highest $AP_{val}$ in addition to the mean and standard deviation of $AP_{val}$ across the 5 runs.

\begin{table*}[]
\centering
\caption{Experiment 1: Binary Classification  Results.}
\resizebox{0.9\width}{!}{
\begin{tabular}{||l|c|c|c|c|c|c|c|c||}
 \hline
                & \multicolumn{4}{|c|}{Character frame}    & \multicolumn{4}{|c||}{Motion}                \\ \hline
method          & model   & agg  & $AP_{val}$  & $AP_{test}$ & model      & agg  & $AP_{val}$      & $AP_{test}$ \\ \hline
                           random &          - &       - &          0.094 &  0.119 &          - &         - &          0.096 &  0.122 \\ \hline
$h_1$ &          - &       - &          0.006 &  0.017 &          - &         - &              - &      - \\ 
$h_2$ &          - &       - &          0.177 &  0.207 &          - &         - &              - &      - \\
$h_3$ &          - &       - &         0.234 & {\bf 0.248} &          - &         - &              - &      - \\
$h_4$ &          - &       - &              - &      - &          - &         - &          0.192 & {\bf 0.163} \\ 
$h_5$ &          - &       - &              - &      - &          - &         - &          0.134 &  0.132 \\  \hline
CLIP  &        $\text{MLP}_M$ &    mean &  0.253$\pm$0.023 &   0.240 &  $\text{MLP}_L$ &    cat &   0.131$\pm$0.010 &  0.107 \\ 
RN50  &  $\text{MLP}_L$ &  cat &  0.266$\pm$0.015 &  0.269 &        $\text{MLP}_M$ &      mean &    0.120$\pm$0.010 &  0.136 \\ 
EN7   &    XGB &    mean &  0.261$\pm$0.025 & {\bf 0.352} &  $\text{MLP}_L$ &      mean &  0.118$\pm$0.007 &  0.129 \\ 
R(2+1)D &  $\text{MLP}_L$ &    mean &  0.222$\pm$0.021 &  0.224 &  $\text{MLP}_L$ &    cat &  0.184$\pm$0.022 &  {\bf 0.193} \\ 
Swin  &        $\text{MLP}_M$ &    mean &    0.270$\pm$0.030 &  0.287 &  $\text{MLP}_S$ &      mean &  0.155$\pm$0.016 &   0.150 \\ 
C4C   &  $\text{MLP}_L$ &    mean &  0.277$\pm$0.012 &  0.304 &        $\text{MLP}_M$ &      mean &  0.121$\pm$0.007 &   0.130 \\ 
YN    &    XGB &  cat &  0.174$\pm$0.015 &   0.160 &  $\text{MLP}_L$ &  diff &  0.136$\pm$0.016 &  0.136 \\ \hline 
YN-CLIP &    XGB &  cat &  0.245$\pm$0.022 &  0.286 &    XGB &  diff &  0.137$\pm$0.016 &  0.137 \\ 
YN-RN50 &    XGB &  cat &  0.283$\pm$0.015 &  0.321 &    XGB &      mean &  0.129$\pm$0.007 &  0.145 \\ 
YN-EN7 &    XGB &    mean &   0.275$\pm$0.030 &  {\bf 0.355} &    XGB &  diff &  0.136$\pm$0.011 &  0.128 \\ 
YN-R(2+1)D &    XGB &  cat &   0.219$\pm$0.020 &  0.218 &    XGB &      mean &   0.170$\pm$0.015 & {\bf 0.177} \\
YN-Swin &  $\text{MLP}_L$ &    mean &  0.269$\pm$0.024 &  0.336 &    XGB &      mean &   0.139$\pm$0.010 &  0.161 \\
YN-C4C &    XGB &    mean &  0.289$\pm$0.017 &  0.327 &    XGB &      mean &   0.140$\pm$0.008 &  0.124 \\ \hline
\end{tabular}
}
\label{table:experiment1}
\end{table*}

\subsection{Experiment 1: Binary classification}
In this experiment, we explore pretrained image, video, and audio networks as embedding (i.e. representation) extractors. Once we extract penulatimate layer embeddings for each shot, we train a binary classifier, which we use to report evaluation metrics. Concretely, for each pair of shots $s^m_i$ and $s^m_j$, we extract representations $r^m_i$ and $r^m_j$. These tensors are then aggregated using a function that takes two tensors and outputs a single tensor. The aggregated tensor is used as the input into the binary classification model. We experimented with using one layer prior to the penultimate layer as well as end-to-end training and were not able to produce reasonable results. We hypothesize that this could be due to the relatively small size of our labeled data. 

We experimented with 5 binary classifiers: (1) XGBoost (XGB) \cite{chen2016xgboost} with 100 boosted rounds, unbounded depth, and logistic objective, (2) logistic regression (LR) with $L_2$ penalty, $C=1$, and LBFGS solver, (3) 2-layer multi-layer perceptron ($\text{MLP}_S$) with ReLU activation \cite{agarap2018deep}, Adam optimizer \cite{kingma2014adam}, lr=0.001, 200 max epochs, and 50 units in each hidden layer, (4) $\text{MLP}_M$ same as (3) but with 100 units in each layer, and (5) $\text{MLP}_L$ same as (3) but with 500 units. We use the XGBoost \cite{chen2016xgboost} library for (1) and scikit-learn \cite{pedregosa2011scikit} for (2-5). We also explored 3 aggregation functions (agg): concatenation (cat), mean pooling (mean), and pairwise absolute distance (diff).

For each encoder and type we train all 15 combinations of models and aggregation functions as described earlier. We then report the combination with the highest mean $AP_{val}$.


\subsubsection{Encoders}

We used a total of seven image, video, and audio encoders. For image encoders, we use CLIP \cite{openai-clip}, ResNet-50 (RN50) \cite{resnet}, and EfficientNet-B7 (EN7) \cite{tan2019efficientnet}. For RN50 and EN7 we use the PyTorch \cite{paszke2019pytorch} implementation pretrained on ImageNet \cite{krizhevsky2012imagenet}. We pass the center frame through the network and extract the penultimate layer embeddings.

For video encoders, we use Video Swin \cite{liu2021video}, R(2+1)D \cite{r2plus1d}, and CLIP4Clip (C4C)  \cite{CLIP4Clip}. For Video Swin, we uniformly sample four views of 32 frames with stride 2 in the temporal dimension. For R(2+1)D, we sample every four frames as described in Sec. \ref{sec:heuristics}. For C4C, we first extract one frame per second and then uniformly sample a maximum of 100 frames. For Video Swin, we use the official implementation \cite{videoswin} pretrained on Kinetics-600 \cite{kinetics600}. For R(2+1)D, we use the PyTorch \cite{paszke2019pytorch} implementation pretrained on Kinetics-400 \cite{kinetics400}, and for C4C we use the CLIP ViT-B/32 encoder trained on MSR-VTT \cite{xu2016msr-vtt}.

Although frame and motion match cutting are largely visual tasks, we wondered whether audio carries any signal that can be used for matching. Therefore, we used YAMNet (YN) \cite{tensorflowmodelgarden2020}, pretrained on the AudioSet dataset \cite{gemmeke2017audio} by feeding 16 kHz mono audio from the video and taking the average over the 1024-dimensional embeddings produced for each 0.48 seconds.

Finally, we also consider the concatenation of each image and video encoder with YN. These audio-visual encoders YN-X are constructed by concatenating the YN embeddings with those for X, where X is either an image encoder or a video encoder (e.g. YN-CLIP).

\subsubsection{Experiment 1 Results}

We present the results in Table \ref{table:experiment1}. 
We use a simple random baseline method that assigns a random score between 0 and 1 for each candidate pair. The expected $AP$ of this random predictor represents the positive prevalence rate (see more details in Supplementary \ref{supp:evaluation}). We include the five heuristics in the table to serve as an additional baseline.

For character frame matching, all visual encoders outperform the random predictor and heuristics $h_1$ and $h_2$, while $h_3$ outperforms CLIP and R(2+1)D. The compound scaling methodology in EN7 \cite{tan2019efficientnet} tends to produce representations that capture additional object details, which may explain why it performs best for this task (see Supplementary \ref{supp:annotation} for a description of the nuanced criteria for this task).

Audio (YN) doesn't perform well in isolation, but audio-visual encoders outperform their visual-only counterparts in most cases. Anecdotally, we have observed cases where  matching pairs have similar background music, but this pattern is not very consistent. 

Motion match cutting is a more challenging and subjective task relative to character frame (see Sec. \ref{annotations}). All methods beat the random predictor, video encoders perform better than image encoders, and R(2+1)D achieves the best results amongst them. This makes sense since video encoders are capable of producing representations that capture motion. Also, both $h_4$ and $h_5$ do well, as the underlying models, Farnebäck and RAFT, were both specifically designed to capture motion via optical flow. Audio doesn't appear to carry any additional signal for this task.

For both match types, we are able to train models with improved retrieval quality over the heuristic-based approaches. The downside is that we cannot easily use a classifier with ANN methods, as discussed in Sec. \ref{scalability}. In the next experiment we explore the possibility of retaining the same level of retrieval quality while using a method that is amenable to ANN methods.

\subsection{Experiment 2: Metric Learning}

Experiment 1 demonstrated that binary classification is an improvement over heuristics, but is not scalable for inter-movie matching. A suitable approach to achieve this is metric learning, which maps a feature vector to a new embedding vector that can be directly searched through using nearest neighbor methods (rather than running a classifier for each pair). Here we take fixed shot encodings from Experiment 1, and learn a new embedding space via noise contrastive estimation \cite{nce}.
We leverage the \texttt{pytorch-metric-learning} library \cite{musgrave2020pytorch}, with NTXentLoss (a.k.a. InfoNCE) \cite{NIPS2016_6b180037}, \cite{infonce} and TripletMarginMiner with hard triplets and cosine similarity. The transformed embeddings can be indexed and retrieved with ANN methods as discussed in Sec. \ref{scalability}.

We experiment with embeddings from three image encoders: CLIP \cite{openai-clip}, RN50 \cite{resnet}, and EN7 \cite{tan2019efficientnet}, three video encoders: R(2+1)D \cite{r2plus1d}, Swin \cite{liu2021video}, and C4C \cite{CLIP4Clip}, and one audio encoder: YN \cite{tensorflowmodelgarden2020}. For both character frame and motion matching, we use a 2-layer MLP with leaky ReLU \cite{xu2015empirical} activation and train with Adam \cite{kingma2014adam} optimizer. A set of hyperparameters are separately tuned for frame matching and motion matching datasets and are used across experiments of all encoders. More details for this experiment can be found in Supplementary \ref{supp:baseline}.

\begin{table}[]
\centering
\caption{Experiment 2: Metric Learning Results.}
\resizebox{\columnwidth}{!}{
\begin{tabular}{||c|cc|cc||}
 \hline
 & \multicolumn{2}{|c|}{Character frame} & \multicolumn{2}{|c||}{Motion} \\ \hline
Encoder         & {\it $AP_{val}$ }  & {\it $AP_{test}$} & {\it $AP_{val}$} & {\it $AP_{test}$ } \\ \hline
CLIP            & 0.231$\pm$0.009      & 0.321         & 0.112$\pm$0.001       & 0.138       \\ 
RN50       & 0.258$\pm$0.010       & 0.343         & 0.121$\pm$0.005       & 0.133       \\ 
EN7 &  0.283$\pm$0.010 & {\bf 0.373}   & 0.139$\pm$0.005       & 0.132       \\ 
R(2+1)D         & 0.219$\pm$0.006      & 0.261         & 0.173$\pm$0.001 & {\bf 0.217} \\ 
Swin      & 0.255$\pm$0.006      & 0.363         & 0.143$\pm$0.002       & 0.170        \\ 
C4C       & 0.273$\pm$0.010       & 0.360          & 0.119$\pm$0.001       & 0.144       \\ 
YN          & 0.115$\pm$0.010       & 0.102         & 0.122$\pm$0.001       &0.124        \\ \hline
\end{tabular}
}
\label{table:metric_learning1}
\end{table}
The results are shown in Table \ref{table:metric_learning1}. For both frame and motion matching, we find that metric learning is able to achieve higher retrieval quality relative to binary classification, while providing superior inference scalability. EN7 and R(2+1)D perform the best for character frame and motion matching (consistent with the previous experiment).
\section{Limitations and Discussion}

%
One limitation of our work is that we have only addressed two specific types of match cutting. We hope to extend this work to other types of match cuts, such as action, sound, light, subject matter, or a broad generalization.

Although our heuristics were designed to aid annotation efforts and are not our main contribution, we will discuss some limitations with them. For frame matching, heuristics 1 was designed to be a crude heuristic and suffers from a lack of spatial alignment between faces. Heuristics 2 and 3 are fairly robust to character framing, though we observe that they can struggle with closeups of faces, where matching face key points may be more suitable. Heuristic 4 and 5 for motion matching  tend to surface shots with similar camera movement (as opposed to action). This occurs because the background covers more pixel area than the foreground in most shots. We hope to address this in the future with foreground-background segmentation. Also, our choice of averaging of optical flow outputs over multiple frames discards temporal information.

Furthermore, our proposed frame matching system is limited to only human characters. Non-human objects and shapes can also make for very interesting matches. Lastly, we acknowledge that there is a domain gap between the datasets the models in our experiments were pretrained on and Hollywood produced movies.



\section{Conclusions}
We have presented a system to take the task of generating good match cut candidates from a manual process akin to finding a needle in a haystack to a semi-automated process that finds "cuts that work" in a fraction of the time.

We summarize our contributions as follows: (1) we present a new and previously unaddressed problem of finding specific match cut types, along with a novel system to identify good candidates. (2) We release a dataset of $\sim$20k annotated shot pairs for two types: character frame and motion matching. (3) We conduct experiments to evaluate our system using the collected dataset. (4) We release code and embeddings for reproducing our experiments.

While we have presented the first system for finding match cuts, there is plenty of room for improvement. Match cutting is a very challenging task relative to coarse-grained object and action recognition tasks and requires more sophisticated video understanding methods that can capture the nuances.

\newpage

\section{Supplemental material}

\subsection{Annotation task details and dataset statistics}\label{supp:annotation}

In this section we describe the rules developed with our in-house editors for the annotation of match cuts,
including examples of match cuts that violate or follow the rules. We show the user interface for annotation,
then provide some additional data set statistics.

\subsubsection{Character frame matching}
\textbf{Rules}

\begin{enumerate}
    \item Proportions and scales of the characters should be the same.
    \item Character poses should be similar.
    \item Shots of the same person are okay, as long as there is something different about the shots. E.g. different location, clothes, time of day.
    \item Shots should not be too similar.
    \item Matches should be between characters, not objects.
\end{enumerate}

\textbf{Examples}

\begin{table}[H]
\centering
\begin{tabular}{ccc}
\hline
\textbf{Example} & \textbf{Match} & \textbf{Violates} \\ \hline
\includegraphics[width=0.25\textwidth]{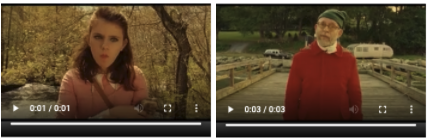} & No & 1 \\ \hline
\includegraphics[width=0.25\textwidth]{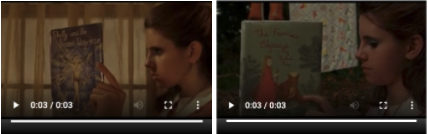} & Yes & None \\ \hline
\includegraphics[width=0.25\textwidth]{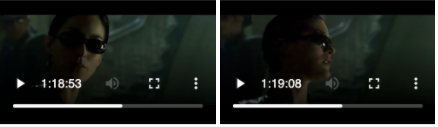} & No & 2, 3, 4 \\
\end{tabular}
\end{table}

Examples are from Moonrise Kingdom \cite{anderson_2012} and The Matrix \cite{wachowski_1999}.

\subsubsection{Motion Match Cutting}

\textbf{Rules}

\begin{enumerate}
    \item Characters/objects should be moving the same way or the camera motion should be similar. E.g. the camera moves the same direction, or an action-reaction pair in which they move opposite directions.
    \item Number of subjects does not have to be the same, as long as the movement, pace and direction are similar.
    \item Shots should not be blurry even if the motion is matching.
\end{enumerate}

\textbf{Examples}

\begin{table}[H]
\centering
\begin{tabular}{ccc}
\hline
\textbf{Example} & \textbf{Match?} & \textbf{Violates} \\ \hline
\includegraphics[width=0.25\textwidth]{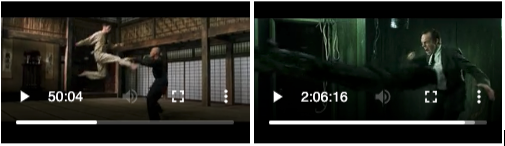} & Yes & None \\ \hline
\includegraphics[width=0.25\textwidth]{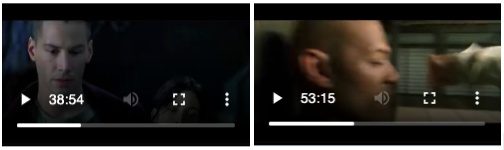} & No & 3 \\ 
\end{tabular}
\end{table}

Examples in this table are from The Matrix \cite{wachowski_1999}.

\subsubsection{User interface for annotation}
We built a custom application for presenting pairs of shots to annotators and collecting labels.
\begin{figure}[H]
    \centering
    \includegraphics[width=1\linewidth]{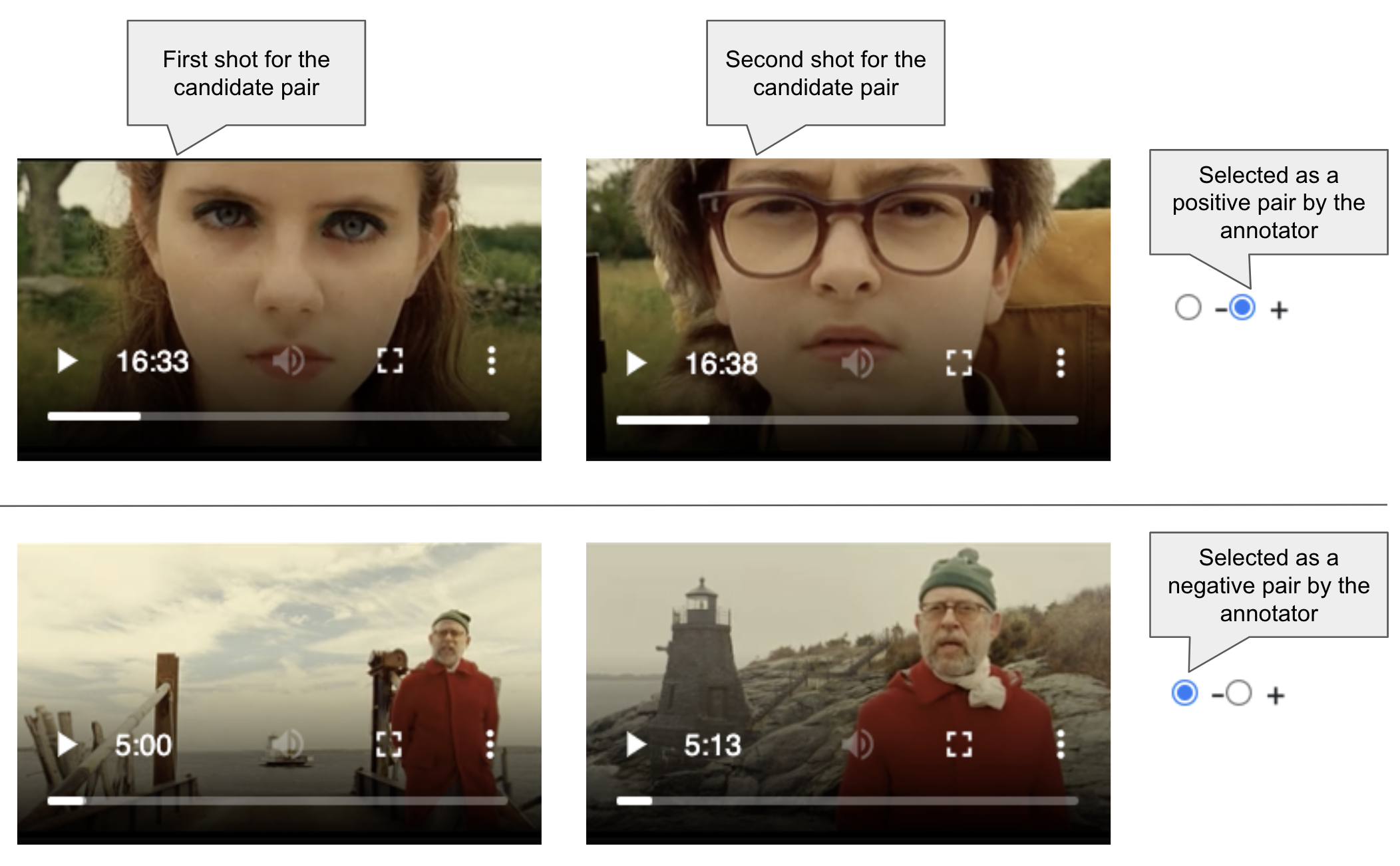}
\end{figure}

Examples are from Moonrise Kingdom (2012) \cite{anderson_2012}.

\subsubsection{Dataset statistics}
\begin{table}[h]
\centering
\begin{tabular}{llll}
\hline
\textbf{Task}                   & \textbf{Frame} & \textbf{Motion} & \textbf{Overall} \\ \hline
Annotated pairs                 & 9,985          & 9,320           & 19,305           \\
Positive pairs (majority)       & 867            & 927             & 1,794            \\
Positive rate                   & 0.087          & 0.099           & 0.093            \\
Pairs w/perfect agreement       & 8,373          & 7,027           & 15,400           \\
Perfect agreement rate          & 0.839          & 0.754           & 0.798            \\
\end{tabular}
\end{table}

\subsubsection{Heuristic positive rate}
\begin{table}[H]
\centering
\begin{tabular}{llll}
\hline
\textbf{Heuristic} & \textbf{Positive pairs} & \textbf{Positive rate} \\ \hline
$h_1$              & 69              & 0.014                  \\
$h_2$              & 808             & 0.162                  \\
$h_4$              & 543             & 0.109                  \\
$h_5$              & 494             & 0.099                  \\
\end{tabular}
\end{table}

Each heuristic selected 5,000 pairs.

\subsubsection{Annotator-level agreement by task}
\begin{figure}[H]
    \centering
    \includegraphics[width=1\linewidth]{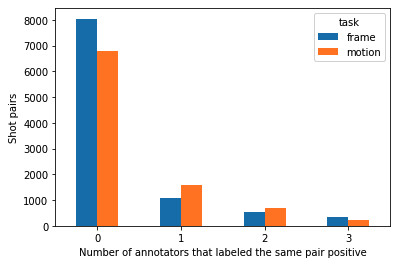}
\end{figure}

\subsubsection{Annotation candidate pair generation}
\textbf{High-level process}
\begin{figure}[H]
    \centering
    \includegraphics[width=1\linewidth]{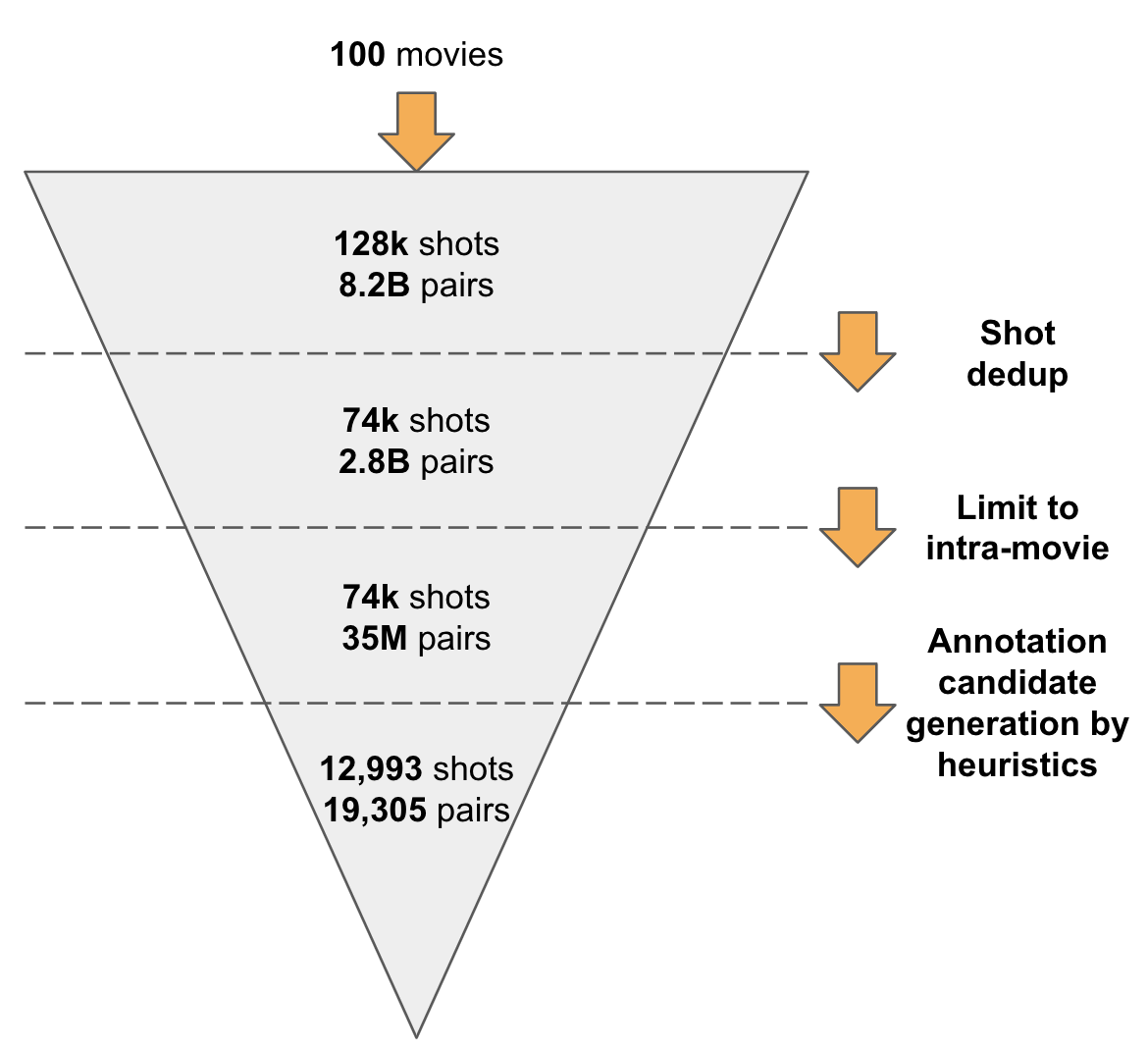}
\end{figure}

\textbf{Statistics}


\begin{table}[H]
\centering
\begin{tabular}{lll}
\hline
\textbf{Stage} & \textbf{Shots} & \textbf{Shot pairs} \\ \hline
After shot segmentation & 128,202 & 8,217,812,301 \\
After dedup & 74,493 & 2,774,566,278 \\
Limit to intra-movie & 74,493 & 34,554,612 \\
Annotated & 12,993 & 19,305
\end{tabular}
\end{table}

\subsection{Title set and shot statistics for the released dataset}\label{supp:title}

\subsubsection{Titles}
The title set is available in the \href{http://github.com/netflix/matchcut}{Github repo}.

\subsubsection{Genre breakdown}
Note that titles can have more than one genre.
\begin{figure}[H]
    \centering
    \includegraphics[width=.92\linewidth]{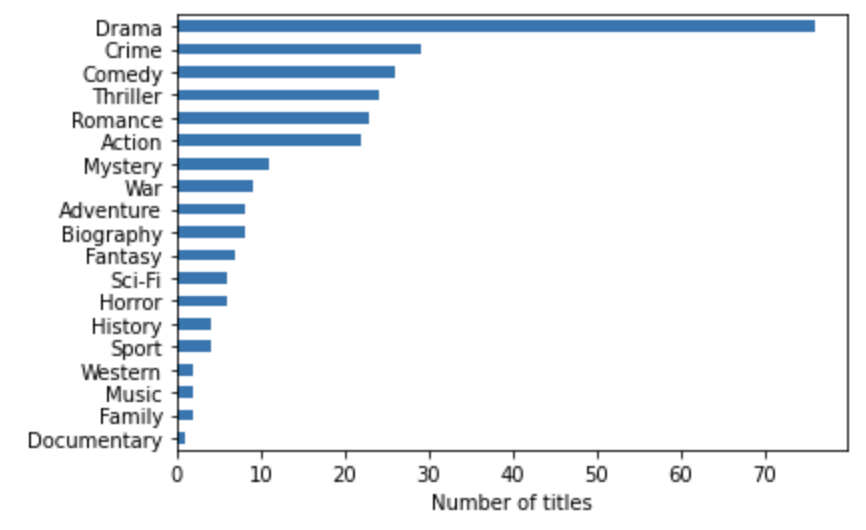}
\end{figure}

\subsubsection{Country breakdown}
\begin{figure}[H]
    \centering
    \includegraphics[width=1\linewidth]{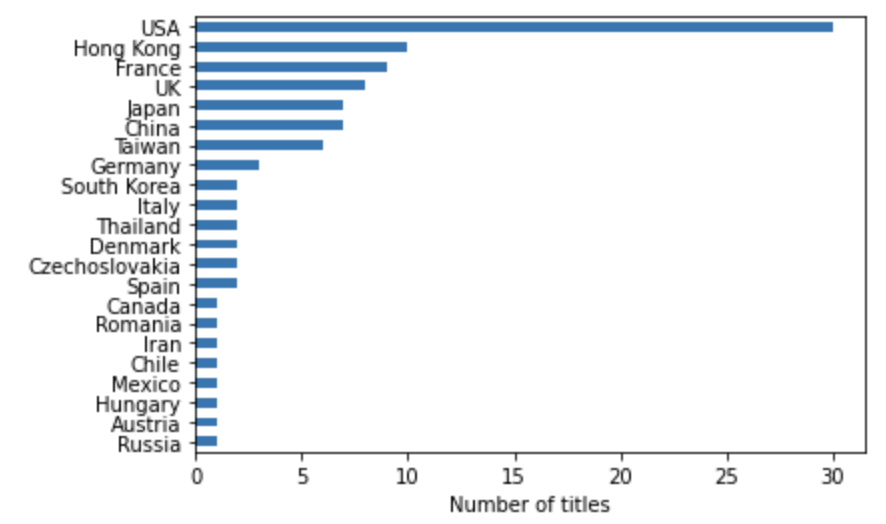}
\end{figure}

\subsubsection{Release year}
\begin{figure}[H]
    \centering
    \includegraphics[width=1\linewidth]{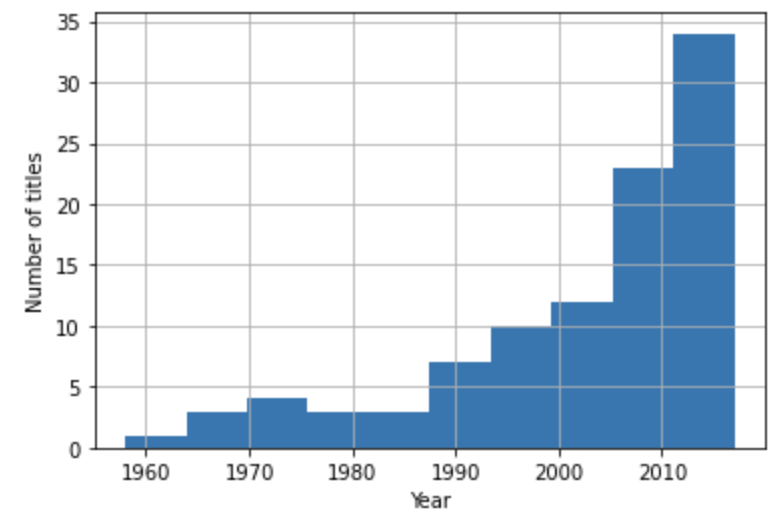}
\end{figure}

\subsubsection{Shot duration statistics}
The duration values are in seconds. Note that these values are computed for the subset of shots that we are releasing (not the entire set of shots in all the titles that we have considered).

\begin{table}[H]
\centering
\begin{tabular}{ll}
\hline
\textbf{Statistic} & \textbf{Value} \\ \hline
Count & 21,205 \\
Mean & 8.174 \\
Std & 15.136 \\
Min & 0.240 \\
25\% & 2.083 \\
25\% & 3.879 \\
75\% & 8.091 \\
Max & 384.500
\end{tabular}
\end{table}

\subsubsection{Shot duration distribution by genre}
Note that these values are computed for the subset of shots that we are releasing (not the entire set of shots in all the titles that we have considered).
\begin{figure}[H]
    \centering
    \includegraphics[width=.9\linewidth]{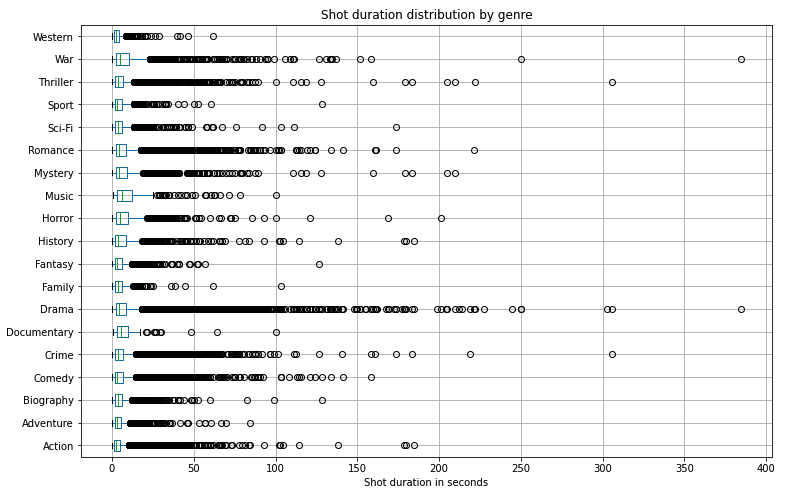}
\end{figure}

\subsubsection{Number of unique shots by title}
Note that these values are computed for the subset of shots that we are releasing (not the entire set of shots in all the titles that we have considered).
\begin{figure}[H]
    \centering
    \includegraphics[width=1\linewidth]{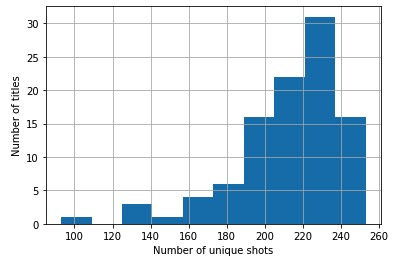}
\end{figure}

\subsection{Evaluation}\label{supp:evaluation}

\subsubsection{Average Precision ($AP$)}
For match cutting, we surface a ranked list of pairs to editors. Ideally, the best candidates should be placed at the top of this list. Average Precision ($AP$) is an information retrieval metric that captures this setup. $AP$ ranges between 0 and 1, where a higher value reflects a higher quality of retrieval.

To demonstrate how $AP$ is calculated in our context, consider the following toy dataset with three labeled pairs (all pairs are from \cite{anderson_2012}):

\begin{table}[H]
\centering
\begin{tabular}{ccc}
\hline
\textbf{Pair} & \textbf{Match} & \textbf{ID} \\ \hline
\includegraphics[width=0.3\textwidth]{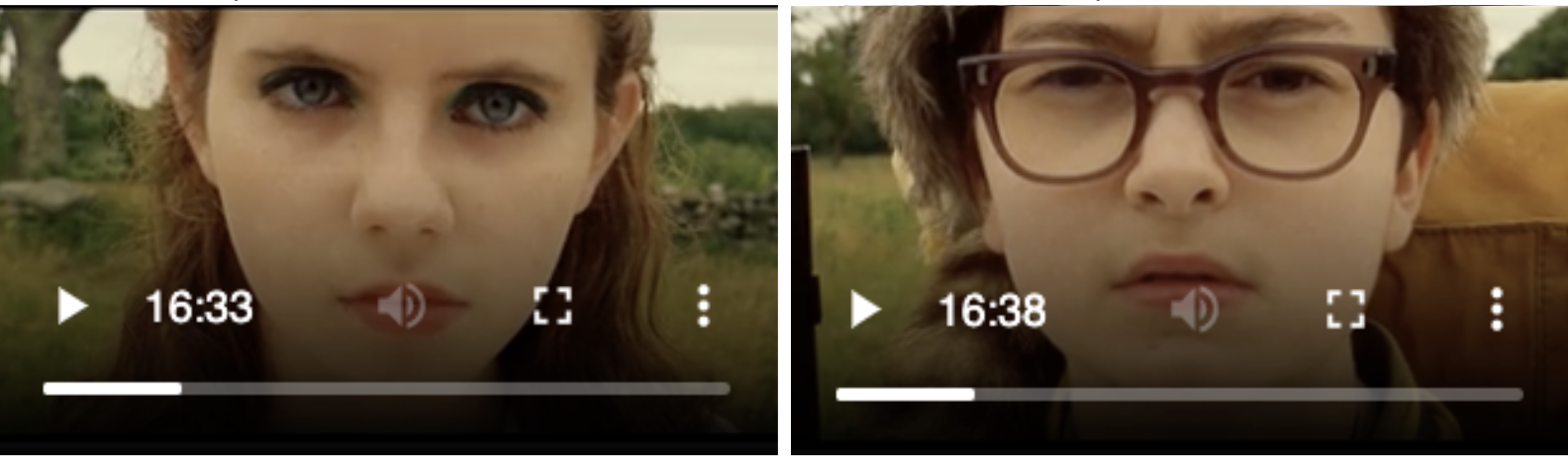} & Yes & A \\ \hline
\includegraphics[width=0.3\textwidth]{supp/frame-scale.png} & No & B \\ \hline
\includegraphics[width=0.3\textwidth]{supp/frame-same-person.png} & Yes & C \\
\end{tabular}
\end{table}

$AP = 1$ is achieved when scores for the positives pairs (i.e. A and C), are higher than the score for the negative pair. For instance, if the scores are 0.9, 0.1, and 0.8 for A, B, and C respectively, then we have $AP=1$. (In this case, the list above would be reordered as A, C, B before it was presented to the editors.)

$AP$ drops below 1 as the scores cause more negatives to be interleaved with positives. For instance, if the scores are 0.9, 0.8, and 0.7 for A, B, and C respectively, then we have $AP=0.83$.

We use the implementation provided by \texttt{scikit-learn} \cite{pedregosa2011scikit}. The following Python snippet shows how $AP$ is calculated for these two cases:
\begin{verbatim}
from sklearn import metrics
ap = metrics.average_precision_score

# after sorting by score we compute 
# precision at each depth
# if the instance is positive and then 
# divide by the number of positives
assert ap(
    y_true=[True, False, True],
    y_score=[0.9, 0.1, 0.8],
) == (1 + 1) / 2
assert ap(
    y_true=[True, False, True],
    y_score=[0.9, 0.8, 0.7],
) == (1 + 2 / 3) / 2
\end{verbatim}

\subsection{Baseline}\label{supp:baseline}
Unlike some metrics such as the Area Under the Receiver Operating Characteristic curve (AUROC), $AP$ is not agnostic to the prevalence of the positive examples (we will call this $p$). In other words, we can expect $AUROC=0.5$ for random guessing regardless of the value of $p$, but $AP=p$ (in expectation) if scores are randomly generated.

Since match cutting is a novel task and no open source benchmarks exist, we treat the positive prevalence $p$ as our baseline, and expect our system to achieve $AP > p$.

The following Python snippet demonstrates that the expected value of $AP$ is $p$:
\begin{verbatim}
import numpy as np
from sklearn import metrics
ap = metrics.average_precision_score

def random_ap(
    n: int, p: float,
) -> float:
    """
    n is the number of candidates.
    p is the positive prevalence.
    """
    assert 0 < p < 1
    scores = np.random.rand(n)
    pos = int(round(p * n))
    true = [True] * pos
    true += [False] * (n - pos)
    return ap(true, scores)

def ap_mean(
    n: int, p: float,
    rounds: int, precision: int = 2,
) -> None:
    aps = [
        random_ap(n=n, p=p)
        for _ in range(rounds)
    ]
    return round(
        np.mean(aps),
        precision,
    )
    
assert ap_mean(
    n=10_000, p=0.2, rounds=1_000,
) == 0.2
assert ap_mean(
    n=10_000, p=0.8, rounds=1_000,
) == 0.8
\end{verbatim}

\subsubsection{Heuristics}
All heuristics described in section 3.3 produce a score given a pair of shots, which can be used for evaluation as described in the previous section. These scores can be used in the same way that we use the output score of a classification model. The only difference is that unlike learned models that can be trained with different seeds, there's no similar source of variation for heuristics. Therefore we only report a single value instead of mean and standard deviation.

\subsection{Experiment 2 hyperparameters}\label{supp:experiment2}
For all experiments we used $\texttt{TripletMarginMiner}$ with $\texttt{type\_of\_triplets="hard"}$ and training batch size of 256.
\noindent
For character frame we use 128 and 1024 hidden units for the first and second layer respectively, and for motion we use 256 and 1024 hidden units in the first two layers of MLP.
For character frame we used 300 epochs and for motion we used 100. 

\subsubsection{Tuning ranges of hyperparameters}
For both tasks we used the following tuning ranges:
\begin{table}[h]
\centering
\begin{tabular}{ll}
\hline
\textbf{Hyperparameter} & \textbf{Range} \\ \hline
$\texttt{temperature}$ & $[10^{-3}, 1]$     \\
$\texttt{learning\_rate}$ & $[10^{-4}, 10^{-1}]$     \\
$\texttt{weight\_decay}$ & $[10^{-4}, 10^{-1}]$     \\
\end{tabular}
\end{table}

\subsubsection{Tuned hyperparameters}
\begin{table}[h]
\centering
\begin{tabular}{lll}
\hline
\textbf{Hyperparameter} & \textbf{Character frame} & \textbf{Motion} \\ \hline
$\texttt{temperature}$ & $7.362 \times 10^{-3}$ & $1.3412 \times 10^{-2}$     \\
$\texttt{learning\_rate}$ & $3.147 \times 10^{-3}$ & $4.056 \times 10^{-4}$     \\
$\texttt{weight\_decay}$ & $10^{-4}$ & $4.54 \times 10^{-4}$     \\
\end{tabular}
\end{table}

\newpage

{\small
\bibliographystyle{ieee_fullname}
\bibliography{egbib}
}

\end{document}